\documentclass{article}

\usepackage[final]{neurips_2021}

\usepackage[utf8]{inputenc} %
\usepackage[T1]{fontenc}    %

\usepackage{hyperref}

\usepackage{url}            %
\usepackage{booktabs}       %
\usepackage{amsfonts}       %
\usepackage{nicefrac}       %
\usepackage{microtype}      %
\usepackage{xcolor}         %
\definecolor{pearDark}{HTML}{2980B9}
\hypersetup{
    colorlinks=true,
    linkcolor=blue,
    filecolor=magenta,      
    urlcolor=blue,
    citecolor=blue
}
\usepackage{graphicx}
\usepackage{float}
\usepackage{multirow}
\usepackage{wrapfig}
\usepackage{mathtools}
\usepackage{caption}

\newcommand{\pargraphSP}{\vspace{-10pt}}

\title{Overcoming the Domain Gap in Contrastive Learning of Neural Action Representations}
\newcommand*\samethanks[1][\value{footnote}]{\footnotemark[#1]}
\author{%
  Semih G\" unel\hspace{-1pt}$^{~1,2}$, Florian Aymanns$^{\:2}$, Sina Honari $^{1}$, Pavan Ramdya \thanks{Equal contribution} $^{\,2}$, Pascal Fua \samethanks~$^{\,1}$
  \AND
  $^{1}$\texttt{CVLab, EPFL,} \texttt{\{name.surname\}@epfl.ch} \\
  $^{2}$\texttt{Neuroengineering Lab, EPFL,} \texttt{\{name.surname\}@epfl.ch} \\
}

\begin{document}

\vspace{-10pt}
\maketitle

\vspace{-10pt}
\begin{abstract}

A fundamental goal in neuroscience is to understand the relationship between neural activity and behavior. For example, the ability to extract behavioral intentions from neural data, or neural decoding, is critical for developing effective brain machine interfaces. Although simple linear models have been applied to this challenge, they cannot identify important non-linear relationships. Thus, a self-supervised means of identifying non-linear relationships between neural dynamics and behavior, in order to compute neural representations, remains an important open problem. To address this challenge, we generated a new multimodal dataset consisting of the spontaneous behaviors generated by fruit flies, \textit{Drosophila melanogaster}---a popular model organism in neuroscience research. The dataset includes 3D markerless motion capture data from six camera views of the animal generating spontaneous actions, as well as synchronously acquired two-photon microscope images capturing the activity of descending neuron populations that are thought to drive actions. Standard contrastive learning and unsupervised domain adaptation techniques struggle to learn \textit{neural action representations} (embeddings computed from the neural data describing action labels) due to large inter-animal differences in both neural and behavioral modalities. To overcome this deficiency, we developed simple yet effective augmentations that close the inter-animal domain gap, allowing us to extract behaviorally relevant, yet domain agnostic, information from neural data. This multimodal dataset and our new set of augmentations promise to accelerate the application of self-supervised learning methods in neuroscience.
\end{abstract}

\vspace{-10pt}
\section{Introduction}
\vspace{-5pt}

Recent technological advances have enabled large-scale simultaneous recordings of neural activity and behavior in animals including rodents, macaques, humans and the vinegar fly, \textit{Drosophila melanogaster}~\cite{dombeck, Seelig, chen2018imaging, lfads, Ecker2010c, wirelesshuman}.
In parallel, recent efforts have been made it possible to perform markerless predictions of 2D and 3D animal poses~\cite{leap,Mathis,Gunel,Bala,newell,couzin,fang2017rmpe,wei2016cpm,cao2017realtime,lp3d, li2020deformation}.
Video and pose data have been used to segment and cluster temporally related behavioral information \cite{task_programming, Segalin2020, berman21, quantify, robertZebraFish20}.
To capture a similarly low dimensional representation of neural activity, most efforts have focused on the application of recurrent state space models \cite{Nassar2019TreeStructuredRS, Linderman621540, pmlr-v54-linderman17a}, or variational autoencoders \cite{Gao2016LinearDN, lfads}. By contrast, there has been relatively limited work aimed at extracting behavioral information from neural data \cite{behavenet, subspace, MLfordecoding} and most efforts have focused on identifying linear relationships between these two modalities using simple correlation analysis, or generalized linear models \cite{subtrate, musall19, stringer19}. However, \textit{neural action representations}---the mapping of behavioral information within neural data---which are particularly crucial for brain-machine interfaces and closed-loop experimentation \cite{bmi, closed-loop} are highly nonlinear. Therefore, devising a systematic approach for uncovering complex non-linear relationships between behavioral and neural modalities remains an important challenge.

Contrastive learning is one promising approach to address this gap. It has been used to extract information from multimodal datasets in a self-supervised way, for modalities including audio, speech, and optical flow~\cite{munro20multi, Han20, alwassel_2020_xdc, asano2020self, relja18, asano2020labelling}. Contrastive learning also has been applied to unimodal datasets, including the study of human motion sequences \cite{liu2020snce, su2020predict, Lin_2020}, medical imaging \cite{chaitanya2020contrastive, zhang2020contrastive}, video understanding \cite{pan2021videomoco, Dave2021TCLRTC}, and pose estimation \cite{honar21SSL, mitra2020multiview}. Thus, contrastive learning holds great promise for application in neuroscience.

 One of the largest barriers to applying contrastive learning to behavioral-neural multimodal datasets is the fact that their statistics (e.g., neuron locations and sizes, body part lengths and ranges of motion) often differ dramatically across animals. This makes it difficult to train models that can generalize across subjects. We confront this domain gap when comparing neural imaging datasets from two different flies \textbf{(Supplementary Fig.~\ref{fig:domain}; Supplementary Videos {\color{blue} 1-2})}. Although multimodal domain adaptation methods for downstream tasks such as action recognition exist \cite{munro20multi}, they assume supervision in the form of labeled source data. However, labeling behavioral-neural datasets requires expensive and arduous manual labor by trained scientists, and thus often leaving  the vast majority of data unlabeled. Similarly, it is non-trivial to generalize few-shot domain adaptation methods to multimodal tasks \cite{kangcontrastive, wang2021crossdomain}. Thus, the field of neuroscience needs new computational approaches that can extract information from ever-increasing amounts of unlabeled multimodal datasets that also suffer from extensive domain gaps across subjects.

Here, we address this challenge by extracting domain agnostic action representations from neural data. We measure representation quality using an action recognition task, in which we apply a linear classification head and transfer our pretrained weights to classify action labels. Therefore, we call our representations \textit{neural action representations}. To best reflect real world conditions, during the unsupervised pre-training phase, we assume access to the paired behavioral-neural data for all domains but without any action labels.
Then, we show that a strong domain gap exists across data taken from different animals, rendering standard contrastive methods ineffective. To address this challenge, we propose a set of simple augmentations that can perform domain adaptation and extract useful representations. We find that the resulting model outperforms baseline approaches, including linear models, previous neural representation learning approaches and common domain adaptation techniques. Finally, to accelerate the uptake and development of these and other self-supervised methods in neuroscience, we will release our new multimodal \textit{Drosophila} behavioral-neural dataset along with associated dense action labels for spontaneously-generated behaviors %

\vspace{-15pt}
\section{Methods}
\vspace{-5pt}
\subsection{Problem Definition}
\vspace{-5pt}
We assume a paired set of data $\mathcal{D}_{s}=\left\{\left(\mathbf{b}_{\mathbf{i}}^{s}, \mathbf{n}_{\mathbf{i}}^{s}\right)\right\}_{i=1}^{n_{s}}$, where $\mathbf{b}^{s}_\mathbf{i}$ and $\mathbf{n}^{a}_\mathbf{i}$ represent the behavioral and neural information respectively, with $n_s$ being the number of samples for animal $s\in \mathcal{S}$. 
We quantify behavioral information $\mathbf{b}^{s}_\mathbf{i}$ as a set of 3D poses corresponding to a set of frames $\mathbf{i}$ from animal $s$, and neural information $\mathbf{n}^{s}_\mathbf{i}$ as a set of two-photon microscope images capturing the activity of neurons. We assume that the two modalities are always synchronized (paired), and therefore describe the same set of events. Our goal is to learn a parameterized image encoder function $f_n$, which maps a set of neural images $\mathbf{n}^{s}_\mathbf{i}$ to a low-dimensional representation. We aim for our learned representation to be representative of the underlying behavioral label, while being modality-agnostic and not representative of the underlying animal identity information $s$, and therefore effectively removing the domain gap across animals 
and modalities. We assume that we are not given behavioral labels during unsupervised training. 

\vspace{-10pt}
\subsection{Contrastive Representation Learning}
\vspace{-5pt}
For each input pair $\left(\mathbf{b}_{\mathbf{i}}^{s}, \mathbf{n}_{\mathbf{i}}^{s}\right)$, we first draw a random view 
$( \tilde{\mathbf{b}}_{\mathbf{i}}^{s}, \tilde{\mathbf{n}}_{\mathbf{i}}^{s} ) $ with a sampled transformation function $t_{n} \sim \mathcal{T}_n$ and $t_{b} \sim \mathcal{T}_b$ , where $\mathcal{T}_n$ and $\mathcal{T}_b$ represent a family of stochastic image transformation functions for behavioral and neural data, respectively. Next, the encoder functions $f_b$ and $f_n$ transform input data into low-dimensional vectors $\mathbf{h}_b$ and $\mathbf{h}_n$, followed by non-linear projection functions $g_b$ and $g_n$, which further transform data into the vectors $\mathbf{z}_b$ and $\mathbf{z}_n$. During training, we sample a minibatch of N input pairs $\left(\mathbf{b}_{\mathbf{i}}^{s}, \mathbf{n}_{\mathbf{i}}^{s}\right)$, and train with the symmetric loss function

\begin{equation}
\mathcal{L}_{NCE} = - \sum_{i=1}^{N} \log \frac{\exp \left(\left\langle\mathbf{z}^{i}_{b}, \mathbf{z}^{i}_{n}\right\rangle / \tau\right)}{\sum_{k=1}^{N} \exp \left(\left\langle\mathbf{z}^{i}_{b}, \mathbf{z}^{k}_{n}\right\rangle / \tau\right)} + \log \frac{\exp \left(\left\langle\mathbf{z}^{i}_{n}, \mathbf{z}^{i}_{b}\right\rangle / \tau\right)}{\sum_{k=1}^{N} \exp \left(\left\langle\mathbf{z}^{i}_{n}, \mathbf{z}^{k}_{b}\right\rangle / \tau\right)}
\label{eq:nce}
\end{equation}

 where $\left\langle\mathbf{z}^{i}_{b}, \mathbf{z}^{i}_{n}\right\rangle$ is the cosine similarity between behavioral and neural modalities and $\tau \in \mathbb{R}^{+}$ is the temperature parameter. The loss function maximizes the mutual information between two modalities \cite{oord2019representation}. The symmetric version of the contrastive loss function was previously used in multimodal self-supervised learning \cite{zhang2020contrastive, Yuan_2021_CVPR}. An overview of our method for learning $f_n$ is shown in \textbf{Supplementary Fig~\ref{fig:method}}. Although standard contrastive learning bridges the gap between different modalities, it does not bridge the gap between different animals, a fundamental challenge that we address in this work through augmentations described in the following section.

\pargraphSP
\paragraph{Swapping Augmentation:}
Given a set of consecutive 3D poses $\mathbf{b}^{s}_\mathbf{i}$, for each $k\in\mathbf{i}$, we stochastically replace $\mathbf{b}^{s}_k$ with one of its nearest pose neighbors in the set of domains $\mathcal{D}_{\mathcal{S}/s}$, where $\mathcal{S}$ is the set of all animals. To do so, we first randomly select a domain $\hat{s} \in \mathcal{S}/s$ and define a probability distribution $\mathbf{P}^{\hat{s}}_{\mathbf{b}^{s}_k}$ over the domain $\mathcal{D}_{\hat{s}}$ with respect to $\mathbf{b}^{s}_k$, 

\begin{equation}
\mathbf{P}^{\hat{s}}_{\mathbf{b}^{s}_k} (\mathbf{b}^{\hat{s}}_l) = 
\frac{\exp ( - \| \mathbf{b}^{\hat{s}}_l - \mathbf{b}^{s}_k \|_{2})}{\sum_{ \mathbf{b}^{\hat{s}}_m \in  \mathcal{D}_{\hat{s}} } \exp ( - \| \mathbf{b}^{\hat{s}}_m - \mathbf{b}^{s}_k \|_{2})}.
\end{equation}

We then replace each 3D pose $\mathbf{b}^{s}_k$ by first uniformly sampling a new domain $\hat{s}$, and then sampling from the above distribution $\mathbf{P}^{\hat{s}}_{\mathbf{b}^{s}_k}$, therefore resulting in $\tilde{\mathbf{b}}^{s}_k \sim \mathbf{P}^{\hat{s}}_{\mathbf{b}^{s}_k}$. In practice, we calculate the distribution $\mathbf{P}$ only over the first $\mathbf{N}$ nearest neighbors of $\mathbf{b}^{s}_k$, in order to sample from a distribution of the most similar poses. We empirically set $\mathbf{N}$ to $128$. Swapping augmentation reduces the identity information in the behavioral data without perturbing it to the extent that semantic action information is lost. Each transformed behavioral sample $\tilde{\mathbf{b}}_{\mathbf{i}}^{s}$ is composed of multiple domains.  This forces the behavioral encoding function $f_{b}$ to leave identity information out, therefore merging multiple domains in the latent space. Swapping augmentation is similar to synonym replacement augmentation used in natural language processing \cite{wei-zou-2019-eda}, where randomly selected words in a sentence are replaced by their synonyms.
To the best of our knowledge, we are the first to use swapping augmentation in the context of time-series analysis or for domain adaptation.

\pargraphSP
\paragraph{Neural Calcium Imaging Data Augmentation:}
Our neural data was obtained using two-photon microscopy and  calcium imaging. The resulting images are only a function of the underlying neural activity, and have temporal properties that differ from the true neural activity. For example, calcium signals from a neuron change much more slowly than the neuron's actual firing rate. Consequently, a single neural image $\mathbf{n}_t$ includes decaying information concerning neural activity from the recent past, and thus carries information about previous behaviors. This makes it harder to decode the current behavioral state. We aimed to prevent this overlap of ongoing and previous actions. Specifically, we wanted to teach our network to be invariant with respect to past behavioral information by augmenting the set of possible past actions.  To do this, we generated new data $\tilde{\mathbf{n}}^{s}_\mathbf{i}$, that included previous neural activity $\mathbf{n}^{s}_k$. To mimic calcium indicator decay dynamics, given a neural data sample $\mathbf{n}^{s}_\mathbf{i}$ of multiple frames, we sample a new neural frame $\mathbf{n}^{s}_k$ from the same domain, where $k \notin \mathbf{i}$. We then convolve $\mathbf{n}^{s}_k$ with the temporally decaying calcium convolutional kernel $\mathcal{K}$, therefore creating a set of images from a single frame $\mathbf{n}^{s}_k$, which we then add back to the original data sample $\mathbf{n}^{s}_\mathbf{i}$. This results in $\tilde{\mathbf{n}}^{s}_\mathbf{i} = \mathbf{n}^{s}_\mathbf{i} + \mathcal{K} * \mathbf{n}^{s}_k$ where $*$ denotes the convolutional operation. 
In the Appendix, we explain calcium dynamics and our calculation of the kernel $\mathcal{K}$ in more detail. 

\vspace{-10pt}
\section{Experiments}
\vspace{-5pt}
In this section we introduce a new dataset consisting of \textit{Drosophila melanogaster} neural and behavioral recordings as well as the set of downstream evaluation metrics.

\vspace{-5pt}
\subsection{Dataset}
\vspace{-5pt}
\paragraph{Motion Capture and Two-photon Dataset (MC2P):}
We acquired data from tethered adult female flies, (\textit{Drosophila melanogaster}). This dataset consists of neural activity recorded using a two-photon microscope \cite{chen18} from the axons of descending neurons passing through the animal's cervical connective. It also includes behavioral data recorded using multi-view infrared cameras \textbf{(Supplementary Fig.~\ref{fig:main}; Supplementary Videos {\color{blue} 1-2})}. Specifically, behavioral video data of size $480\times 960$ pixels were acquired at $100$ frames-per-second (fps) using a six circular camera network with the animal at its center.The neural data was recorded using a two-photon microscope, yielding images of $480\times 736$ pixels at $16$ fps.
Eight animals and 133 trials were recorded, resulting in 8.2 hours of recordings with 2,975,000 behavioral and 476,000 neural frames. The dataset includes manual and dense action labels of eight behaviors: \textit{forward walking}, \textit{pushing}, \textit{hindleg grooming},  \textit{abdominal grooming}, \textit{rest}, \textit{foreleg grooming}, \textit{antennal grooming}, and \textit{eye grooming}. We report the statistics of our dataset in \textbf{Supplementary Fig.~\ref{fig:dataset}}. See the Appendix for more details.
\vspace{-5pt}
\subsection{Evaluation}
\vspace{-5pt}
To evaluate our unsupervised pretrained neural encoder $f_{n}$, we froze its parameters and trained a randomly initialized linear classification layer with with SGD. To compare data efficiency, for each setting we evaluated image encoders with $50\%$ and $100\%$ of the data. We report aggregated results over $4$-fold cross-validation evaluations and report the average in each task. We evaluated models on the following tasks:

\pargraphSP
\paragraph{Single-Animal Action Recognition:} We performed action recognition on a single domain by training and testing on the same animal. We repeated the same experiment on each of four animals, and report the mean accuracy. 
\pargraphSP
\paragraph{Multi-Animal Action Recognition:} We evaluated models on their ability to reduce the domain gap. We trained the linear classifier on N-1 animals and tested on the left-out one, leaving each animal out one at a time. 
\pargraphSP
\paragraph{Identity Recognition:} We classified animal identity from among the eight animals. 
We sampled 1000 random data-points uniformly across animals and applied 4-fold cross validation. In the case that the learned representations are domain (subject) invariant, we expect that the linear classifier will not be able to detect the domain of the representations, resulting in a lower identity recognition accuracy.
\vspace{-10pt}
\section{Results}
\vspace{-5pt}
 We present action recognition results from neural imaging data in \textbf{Table~\ref{tab:clsf}} and identity recognition task results in \textbf{Table~\ref{tab:identity_clsf}}. For the supervised baseline, we trained an MLP with manually annotated action labels using cross-entropy loss, with the raw neural data as input, and show the results in the "Raw" section of ~\textbf{Table~\ref{tab:clsf}}. For the "Self-Supervised" section, before using the proposed augmentations, the contrastive method SimCLR performed worse than convolutional and recurrent regression-based methods including the current state-of-art BehaveNet~\cite{behavenet}. Although domain adaptation methods MMD (Maximum Mean Discrepancy) and GRL (Gradient Reversal Layer) close the domain gap and lower identity recognition accuracy, they do not position semantically similar points near one another (\textbf{Supplementary Fig. \ref{fig:tsne}}). As a result, domain adaptation-based methods do not result in significant improvements in the action recognition task. Although regression-based methods suffer less from the domain gap problem, they do not produce as discriminative representations as contrastive learning based methods. The same trend is observed in Table \textbf{Table~\ref{tab:identity_clsf}}. Our proposed set of augmentations close the domain gap, while significantly improving the action recognition baseline for self-supervised methods, for both single-animal and multi-animal tasks. We include detailed information about the baselines in the Appendix.

\begin{wraptable}{r}{10cm}
\vspace{-35pt}
\centering
\footnotesize{

\begin{tabular}{ll|cc|cc}
\toprule
\multicolumn{2}{r|}{\textbf{Tasks $\rightarrow$}} & \multicolumn{2}{c|}{\textbf{Single-Animal $\uparrow$ }} & \multicolumn{2}{c}{\textbf{Multi-Animal $\uparrow$}} \\
\multicolumn{2}{r|}{\textbf{Percentage of Data $\rightarrow$}} & 0.5 & 1.0 & 0.5 & 1.0 \\
\midrule
\multicolumn{2}{l|}{{Random Guess}} &  16.6 & 16.6 & 16.6& 16.6 \\
\midrule
Neural (Linear) & \parbox[t]{1.5mm}{\multirow{2}{*}{\rotatebox[origin=c]{90}{\scriptsize{\textbf{Raw}}}}} & 29.3  & 32.5 & 18.4 & 18.4\\
Neural (MLP) & &  -- & --	& 18.4 & 18.4 \\
\midrule

SimCLR \cite{simclr} &
\parbox[t]{1.5mm}{\multirow{3}{*}{\rotatebox[origin=c]{90}{\scriptsize{\textbf{Self-Supervised}}}}} & 54.3 & 57.6 & 46.9 & 50.6 \\
Regression (Recurr.) & &  53.6& 59.7 & 49.4 & 51.2 \\
Regression (Conv.) & & 52.6 & 59.6 & 50.6 & 55.8 \\
BehaveNet \cite{behavenet} & & 54.6 & 60.2 & 50.5 & 56.8\\
Ours & & \textbf{57.9} & \textbf{63.3} & \textbf{54.8} & \textbf{61.9} \\

\midrule
SimCLR \cite{simclr} + MMD & \parbox[t]{1.5mm}{\multirow{3}{*}{\rotatebox[origin=c]{90}{\scriptsize{\textbf{Domain Ada.}}}}} &  53.6 & 57.8 & 50.1 & 53.1\\
SimCLR \cite{simclr} + GRL & &  53.5 & 56.3 & 49.9 & 52.3 \\
Regression (Conv.) + MMD  & & 54.5 & 60.7 & 52.6 & 55.4\\
Regression (Conv.) + GRL  & & 55.5 & 60.2 & 51.8 & 55.7 \\

\bottomrule
\end{tabular}
}
\caption{\textbf{Action Recognition Accuracy.} Single- and multi-animal action recognition results on the MC2P dataset. Behavioral and Neural MLP results for the single-animal task are removed because single animals often do not have enough labels for every action.}
\label{tab:clsf}
\end{wraptable}

\vspace{-10pt}
\section{Conclusion}
\vspace{-5pt}
We introduced an unsupervised \textit{neural action representation} framework. We extended previous methods by establishing set of augmentations that we show overcomes the multimodal domain gap in our \textit{Drosophila} behavioral-neural dataset. Finally, we will share in order to dataset to accelerate the application of self-supervised learning methods in neuroscience. In future work, we aim to extend our work for domain generalization.

\newpage

\bibliography{main.bbl}

\section*{Checklist}

\begin{enumerate}

\item For all authors...
\begin{enumerate}
  \item Do the main claims made in the abstract and introduction accurately reflect the paper's contributions and scope?
    \answerYes{}
  \item Did you describe the limitations of your work?
    \answerYes{Please see the Conclusion section.}
  \item Did you discuss any potential negative societal impacts of your work?
    \answerYes{Please see the Broader Impact Statement Section.}
  \item Have you read the ethics review guidelines and ensured that your paper conforms to them?
    \answerYes{}
\end{enumerate}

\item If you are including theoretical results...
\begin{enumerate}
  \item Did you state the full set of assumptions of all theoretical results?
    \answerNA{}
	\item Did you include complete proofs of all theoretical results?
    \answerNA{}
\end{enumerate}

\item If you ran experiments...
\begin{enumerate}
  \item Did you include the code, data, and instructions needed to reproduce the main experimental results (either in the supplemental material or as a URL)?
    \answerYes{We include instructions to download and use our dataset in the supplementary materials.}
  \item Did you specify all the training details (e.g., data splits, hyperparameters, how they were chosen)?
    \answerYes{Please see the appendix, particularly the implementation details section.}
	\item Did you report error bars (e.g., with respect to the random seed after running experiments multiple times)?
    \answerYes{We use cross-validation and report the mean accuracy. Please see the appendix, the implementation details section.}
	\item Did you include the total amount of compute and the type of resources used (e.g., type of GPUs, internal cluster, or cloud provider)?
    \answerYes{Please see the appendix, the implementation details section.}
\end{enumerate}

\item If you are using existing assets (e.g., code, data, models) or curating/releasing new assets...
\begin{enumerate}
  \item If your work uses existing assets, did you cite the creators?
    \answerNA{}
  \item Did you mention the license of the assets?
    \answerYes{We include the license of our dataset in the supplementary material.}
  \item Did you include any new assets either in the supplemental material or as a URL?
    \answerYes{We include instructions to download and using our dataset in the supplementary materials.}
  \item Did you discuss whether and how consent was obtained from people whose data you're using/curating?
    \answerNA{}
  \item Did you discuss whether the data you are using/curating contains personally identifiable information or offensive content?
    \answerNA{}{}
\end{enumerate}

\item If you used crowdsourcing or conducted research with human subjects...
\begin{enumerate}
  \item Did you include the full text of instructions given to participants and screenshots, if applicable?
    \answerNA{}
  \item Did you describe any potential participant risks, with links to Institutional Review Board (IRB) approvals, if applicable?
    \answerNA{}
  \item Did you include the estimated hourly wage paid to participants and the total amount spent on participant compensation?
    \answerNA{}
\end{enumerate}

\end{enumerate}

\setcounter{section}{0}
\renewcommand{\thesection}{S.\arabic{section}}
\renewcommand{\thesubsection}{\thesection.\arabic{subsection}}

\newcommand{\beginsupplementary}{%
	\setcounter{table}{0}
	\renewcommand{\thetable}{S\arabic{table}}%
	\setcounter{figure}{0}
	\renewcommand{\thefigure}{S\arabic{figure}}%
}
\appendix

\newpage
\beginsupplementary
{%
	\centering
	\textbf{\Large Appendix for Overcoming the Domain Gap in \\ \vspace{10pt} $\qquad$ Contrastive Learning of Neural Action Representations}
	\vspace{1. em}
}
\setcounter{page}{1}

\section{Implementation Details}

Aside from the augmentations mentioned in the main text, for the image transformation family $\mathcal{T}_n$, we used a sequential application of Poisson noise, Gaussian blur, and color jittering. 
In contrast with recent work on contrastive visual representation learning, we only applied brightness and contrast adjustments in color jittering because neural images have a single channel that measures calcium indicator fluorescence intensity. We did not apply any cropping augmentation, such as cutout, because action representation is often highly localized and non-redundant (e.g., grooming is associated with the activity of a small set of neurons and thus with only a small number of pixels). We did not apply affine transformations since it removes absolute location information, which is essential for associating neural data with behavioral information (e.g., left-turning is associated with higher activity of descending neurons on the right-side of the connective). We applied the same augmentations to each frame in single sample of neural data. 

For the behavior transformation family $\mathcal{T}_b$, we used a sequential application of scaling, shear, and random temporal and spatial dropping. We did not apply rotation and translation augmentations because the animals were tethered (i.e., restrained from moving freely), and their direction and absolute location were kept fixed throughout the experiment. We did not use time warping because neural and behavioral information are temporally linked (e.g., fast walking has different neural representations than slow walking).

For all methods, we initialized the weights of the networks randomly unless otherwise specified. To keep the experiments consistent, we always paired $32$ frames of neural data with $8$ frames of behavioral data. For the neural data, we used a larger time window because the timescale during which dynamic changes occur are smaller. For the paired modalities, we considered data synchronized if their center frames had the same timestamp. We trained contrastive methods for $200$ epochs and set the temperature value $\tau$ to $0.1$. We set the output dimension of $\mathbf{z}_b$ and $\mathbf{z}_n$ to $128$. We used a cosine training schedule with three epochs of warm-up. For non-contrastive methods, we trained for $200$ epochs with a learning rate of $1e-4$, and a weight decay of $1e-5$, using the Adam optimizer \cite{adam}. We ran all experiments using an Intel Core i9-7900X CPU, 32 GB of DDR4 RAM, and a GeForce GTX 1080. Training for a single SimCLR network for 200 epochs took 12 hours. To create train and test splits, we removed two trials from each animal and used them only for testing.
For the domain adaptation methods GRL and MMD, we reformulated the denominator of the contrastive loss function. Given a domain function $dom$ which gives the domain of the data sample, we replaced one side of $L_{NCE}$ in Eq.~\ref{eq:nce} with,

\begin{equation}
\mathcal{L}_{NCE} = -\sum_{i=1}^{N} \log \frac{\exp \left(\left\langle\mathbf{z}^{i}_{b}, \mathbf{z}^{i}_{n}\right\rangle / \tau\right)}{\sum_{k=1}^{N}  \mathbf{1}_{[dom(i) = dom(k)]}  \exp \left(\left\langle\mathbf{z}^{i}_{b}, \mathbf{z}^{k}_{n}\right\rangle / \tau\right)},
\end{equation}

where selective negative sampling prevents forming trivial negative pairs across domains, therefore making it easier to merge multiple domains. Negative pairs formed during contrastive learning try to push away inter-domain pairs, whereas domain adaptation methods try to merge multiple domains to close the domain gap. We found that the training of contrastive and domain adaptation losses together could be quite unstable, unless the above changes were made to the contrastive loss function.

 We used the architecture shown in \textbf{Supplementary Table~\ref{table:encoder}} for the neural image and behavioral pose encoder. Each layer except the final fully-connected layer was followed by Batch Normalization and a ReLU activation function \cite{batchnorm}. For the self-attention mechanism in the behavioral encoder \textbf{(Supplementary Table~\ref{table:encoder})}, we implement Bahdanau attention~\cite{bahdanau}. Given the set of intermediate behavioral representations $S \in \mathbb{R} ^{T \times D}$, we first calculated,
$$
\mathbf{r}=W_{2} \tanh \left(W_{1} S^{\top}\right) \quad \text { and } \quad \mathbf{a}_{i}=-\log \left(\frac{\exp \left(\mathbf{r}_{i}\right)}{\sum_{j} \exp \left(\mathbf{r}_{j}\right)}\right)
$$
where $W_{1}$ and $W_{2}$ are a set of matrices of shape $\mathbb{R}^{12\times D}$ and $\mathbb{R}^{1\times12}$ respectively. $\mathbf{a}_i$ is the assigned score i-th pose in the sequence of motion. Then the final representation is given by $\sum_{i}^{T} \mathbf{a}_i S_{i}$.

\begin{table}[t]
\setlength{\tabcolsep}{2pt}

\begin{minipage}[t]{0.40\hsize}\centering
\centering
\captionsetup{justification=centering}
\caption*{{\bf (a)} First part of the Neural Encoder $f_n$}
\vspace{3pt}
\scriptsize
\begin{tabular}[t]{  l r c c r }
 \toprule
 Layer & \# filters & K & S   & Output  \\  
 \midrule
 input  & 1 &   - &  -             & $T\times 128 \times 128 $  \\ 
 conv1  & 2 & (3,3)   & (1,1)     & $T\times 128 \times 128 $  \\ 
 mp2   & -   & (2,2)   & (2,2)     & $T\times 64 \times 64 $  \\ 
 conv3 & 4 & (3,3)   & (1,1)     & $T\times 64 \times 64 $  \\ 
 mp4   & -   & (2,2)   & (2,2)     & $T\times 32 \times 32 $  \\ 
 conv5 & 8 & (3,3)   & (1,1)     & $T\times 32 \times 32 $  \\ 
 mp6   & -   & (2,2)   & (2,2)     & $T\times 16 \times 16 $  \\
 conv7 & 16 & (3,3)   & (1,1)     & $T\times 16 \times 16 $  \\
 mp8   & -   & (2,2)   & (2,2)     & $T\times 8 \times 8 $  \\
 conv9 & 32 & (3,3)   & (1,1)     & $T\times 8 \times 8 $  \\
 mp10   & -   & (2,2)   & (2,2)     & $T\times 4 \times 4 $  \\
 conv11 & 64 & (3,3)   & (1,1)     & $T\times 4 \times 4 $  \\
 mp12   &  -   & (2,2)   & (2,2)     & $T\times 2 \times 2 $  \\
 fc13   & 128  & (1,1)   & (1,1)     & $T\times 1 \times 1 $  \\
 fc14   & 128  & (1,1)   & (1,1) & $T\times 1 \times  1$  \\
 \bottomrule
\end{tabular}
\normalsize
\end{minipage}
\hfill
\begin{minipage}[t]{0.55\hsize}\centering
\caption*{{\bf (b)} Behavior Encoder $f_b$}
\vspace{3pt}
\scriptsize
\begin{tabular}[t]{  l r r r r r }
 \toprule
 Layer & \# filters & K & S   & Output  \\  
 \midrule
input  & 60 &   - &  -          & $T\times 60 $  \\  
conv1    & 64 & (3) & (1)   & $T \times 64 $  \\  
conv2 & 80 & (3)   & (1)    & $T \times 80 $  \\ 
mp2   & - & (2)   & (2)     & $T / 2 \times 80 $  \\ 
conv2 & 96 & (3)   & (1)    & $T / 2 \times 96 $  \\ 
conv2 & 112 & (3)   & (1)   & $T / 2 \times 112 $  \\ 
conv2 & 128 & (3)   & (1)   & $T / 2 \times 128 $  \\ 
attention6   & - & (1)   & (1)      & $1 \times 128 $  \\
fc7  & 128  & (1)   & (1)   & $1 \times 128$  \\
 \bottomrule
\end{tabular}
\normalsize

\end{minipage}

\normalsize

\vspace{7pt}
\caption{\textbf{Architecture details.} Shown are half of the neural encoder $f_n$ and behavior encoder $f_b$ functions. How these encoders are used is shown in Supplementary Figure~\ref{fig:method}. Neural encoder $f_n$ is followed by 1D convolutions similar to the behavioral encoder $f_b$, by replacing the number of filters. Both encoders produce $128$ dimensional output, while first half of the neural encoder do not downsample on the temporal axis. \textit{mp} denotes a max-pooling layer. Batch Normalization and ReLU activation are added after every convolutional layer. 
}
\label{table:encoder}
\end{table}

\section{Dataset Details} 
Here we provide a more detailed technical explanation of the experimental dataset. Transgenic female \textit{Drosophila melanogaster} flies aged 2-4 days post eclosion were selected for experiments. They were raised on a 12h:12h day, night light cycle and recorded in either the morning or late afternoon Zeitgeber time. Flies expressed both GCaMP6s and tdTomato in all brain neurons targeted by otd-Gal4 ($;\frac{Otd-nls:FLPo (attP40)}{P{20XUAS-IVS-GCaMP6s}attP40};\frac{R57C10-GAL4, tub>GAL80>}{P{w[+mC]=UAS-tdTom.S}3}$).
The fluorescence of GCaMP6s proteins within the neuron increases when it binds to calcium. There is an increase in intracellular calcium when neurons are active and fire action potentials.
Due to the relatively slow release (as opposed to binding) of calcium by  GCaMP6s molecules, the signal decays exponentially.
We also expressed the red fluorescent protein, tdTomato, in the same neurons as an anatomical fiduciary to be used for neural data registration that compensates for image deformations and translations during animal movements.
We recorded neural data using a two-photon microscope (ThorLabs, Germany; Bergamo2) by scanning the cervical connective. This neural tissue serves as a conduit between the brain and ventral nerve cord (VNC) \cite{chen2018imaging}.
The brain-only GCaMP6s expression pattern in combination with restrictions of recording to the cervical connective allowed us to record a large population of descending neuron axons while also being certain that none of the axons arose from ascending neurons in the VNC. Because descending neurons are expected to drive ongoing actions \cite{Cande}, this imaging approach has the added benefit of ensuring that the imaged cells could, in principle, relate to paired behavioral data.

For neural data processing, raw microscope files were first converted into *.tiff files. These data were then synchronized using a custom Python package \cite{aymanns21utils2p}.
We then estimated the motion of the neurons using images acquired on the red (tdTomato) PMT channel.
The first image of the first trial was selected as a reference frame to which all other frames were registered.
For image registration, we estimated the vector field describing the motion between two frames. To do this, we numerically solved the optimization problem in Eq.~\ref{eq:opticalflow}, where $w$ is the motion field, $I_t$ is the image being transformed, $I_r$ is the reference image, and $\Omega$ is the set of all pixel coordinates \cite{chen2018imaging, aymanns21ofco}.
\begin{equation}
    \label{eq:opticalflow}
    \hat{w} = argmin_w \sum_{x\in\Omega} |I_t(x + w(x)) - I_r(x)| - \lambda \sum_{x\in\Omega} || \nabla w(x) ||^2_2
\end{equation}

A smoothness promoting parameter $\lambda$ was empirically set to 800.
We then applied $\hat{w}$ to the green PMT channel (GCaMP6s).
To denoise the motion corrected green signal, we trained a DeepInterpolation network \cite{deepinterpolation} for nine epochs for each fly and applied it to the rest of the frames.
We only used the first 100 frames of each trial and used the first and last trials as validation data.
The batch size was set to 20 and we used 30 frames before and after the current frame as input.
In order to have a direct correlation between pixel intensity and neuronal activity we applied the following transformation to all neural images $\frac{F - F_0}{F_0} \times 100$, where $F_0$ is the baseline fluorescence in the absence of neural activity. To estimate $F_0$, we used the pixel-wise minimum of a moving average of 15 frames.

 We calibrated the camera rig and extracted 3D poses including 38 landmarks from each animal from RGB video data using DeepFly3D~\cite{Gunel}. We then further preprocessed the 3D data by extracting the anchor (thorax-coxa) joints from each leg and and normalizing the range of the data between $ [0, 1] $. Finally, we applied inverse kinematics to convert 3D poses into Euler angles using \cite{Lobato-Rios2021.04.17.440214}. Unlike human action datasets with scripted actions and a uniform distribution over time, our MC2P dataset is more challenging to analyze because it includes spontaneous and unscripted animal actions with heavy-tailed time and action distributions \textbf{(Supplementary Fig.~\ref{fig:dataset})}.

\paragraph{Calcium Dynamics:}
The relationship between the calcium signal $\mathbf{n}_t$ and neural activity $\mathbf{s}_t$ can be modeled as a first-order autoregressive process
$$\mathbf{n}_t=\gamma \mathbf{n}_{t-1}+ \alpha \mathbf{s}_t,$$
where $\mathbf{s}_t$ is a binary variable indicating an event at time $t$ (e.g. the neuron firing an action potential).
The amplitudes $\gamma$ and $\alpha$ determine the rate at which the signal decays and the initial response to an event, respectively. In general, $0 < \gamma < 1 $, therefore resulting in an exponential decay of information pertaining to $\mathbf{s}_t$ to be inside of $\mathbf{n}_t$. A single neural image $\mathbf{n}_t$ includes decaying information from previous neural activity, and hence carry information from past behaviors. For more detailed information on calcium dynamics, see \cite{pnevmatikakis2013bayesian, Rupprecht21}. Assuming no neural firings, $\mathbf{s}_{t}=0$, $\mathbf{n}_{t}$ is given by $\mathbf{n}_{t} = \gamma^{t} \mathbf{n}_{0}$. Therefore, we define the calcium kernel $\mathcal{K}$ as $\mathcal{K}_t = \gamma ^ {t}$. 

\section{Baseline Methods}

We compare our method with a set of baseline methods previously applied on behavioral-neural datasets.

\begin{wraptable}{r}{0.55\textwidth}
\caption{\textbf{Identity Recognition task.} Comparison of neural representation learning approaches on an Identity Recognition task. Smaller values reflect better representations.}
\label{tab:identity_clsf}
\begin{center}
\small
\setlength{\tabcolsep}{0.2em}
\begin{tabular}{lccccc}
\toprule
& Identity Recog. & Identity Recog. & \\
Method  & (0.5, Accuracy) & (1.0, Accuracy)\\
\midrule
Random Guess & 12.5 & 12.5  \\
\midrule
Behavior (Linear) & 88.6 & 89.7 \\
Neural (Linear) & 100.0 & 100.0 \\
\midrule
SimCLR \cite{simclr} & 69.9 & 80.3  \\
Regression (Recurrent) & 89.5 & 91.8  \\
Regression (Convolution) & 88.7 & 92.5  \\
BehaveNet \cite{behavenet} & 80.2 &  83.4  \\
Ours & \textbf{12.5} & \textbf{12.5}  \\
\midrule
SimCLR + MMD \cite{MMD} & 18.4  &  21.2	\\
SimCLR + GRL \cite{GRL} & 16.7 &  19.1	\\
\bottomrule
\end{tabular}
\end{center}
\end{wraptable}

\paragraph{Supervised:}  We trained a single feedforward network with manually annotated action labels using cross-entropy loss, with the raw data as input. We initialized the network weights randomly. We discarded datapoints that did not have associated behavioral labels. For the MLP baseline, we trained a simple three layer MLP with a hidden layer size of 128 neurons with ReLU activation and without Batch Normalization.

\pargraphSP
\paragraph{Regression (Convolutional):} We trained a single fully-convolutional feedforward network for a behavioral reconstruction task, given the set of neural images. We trained with a simple MSE loss. To keep the architectures consistent, the average pooling was followed by a projection layer. We took the input to the projection layer as the final representation.

\pargraphSP
\paragraph{Regression (Recurrent):} Similar to convolutional regression, the last projection network was replaced with a two-layer GRU module. The GRU module takes as an input the fixed representation of neural images.
At each time step, the GRU module predicts a single 3D pose with a total of eight steps to predict the eight poses associated with an input neural image. We trained this model with a simple MSE loss. We took the input of the GRU module as the final representation of neural encoder. 

\pargraphSP
\paragraph{BehaveNet:} BehaveNet uses a discrete autoregressive hidden Markov model (ARHMM) module to decompose 3D motion information into discrete “behavioral syllables." Similar to regression baseline, the neural information is used to predict the posterior probability of observing each discrete syllable \cite{behavenet}. Unlike the original method, we used 3D poses instead of RGB videos. We skipped compressing the behavioral data using a convolutional autoencoder because, unlike RGB videos, 3D poses are already low-dimensional.

\pargraphSP
\paragraph{SimCLR:} We trained the original SimCLR module without the calcium imaging data and swapping augmentations. Similar to our method, we took the features before the projection layer as the final representation \cite{simclr}.

\pargraphSP
\paragraph{Gradient Reversal Layer (GRL):} Together with the contrastive loss, we trained a two-layer MLP domain discriminator per modality, $D_{b}$ and $D_{n}$, which estimated the domain of the neural and behavioral representations \cite{GRL}. Discriminators were trained with the loss function

$$
\begin{array}{r}
\mathcal{L}_{D}=\sum_{x \in\{\mathbf{b}, \mathbf{n}\}}-d \log \left(D_{m}\left(f_{m}(x)\right)\right) \\
\end{array},
$$

where $d$ is the one-hot identity vector. Gradient Reversal layer is inserted before the projection layer. Given the reversed gradients, the neural and behavioral encoders $f_n$ and $f_{b}$ learn to fool the discriminator and outputs invariant representations across domains. We kept the hyperparameters of the discriminator the same as in previous work \cite{munro20multi}. We froze the weights of the discriminator for the first 10 epochs, and trained only the $\mathcal{L}_{NCE}$. We trained the network using both loss functions, $\mathcal{L}_{NCE} + \lambda_{D} \mathcal{L}_{D}$, for the remainder of training. We set the hyperparameters $\lambda_{D}$ to $10$ empirically.

\pargraphSP
\paragraph{Maximum Mean Discrepancy (MMD):} We replaced adversarial loss with a statistical test to minimize the distributional discrepancy from different domains \cite{MMD}. Similar to previous work, we applied MMD only on the representations before the projection layer independently on both modalities \cite{munro20multi, kangcontrastive}. Similar to the GLR baseline, we first trained 10 epochs only using the contrastive loss, and trained using the combined losses  $\mathcal{L}_{NCE} + \lambda_{MMD} \mathcal{L}_{MMD}$ for the remainder.  We set the hyperparameters $\lambda_{MMD}$ as $1$ empirically.

\section{Broader Impact}
In this work, we have proposed a method that extracts behavioral information from two-photon neural imaging data. In the long run, our work can impact humans through the development of more effective brain machine interface neural decoding algorithms. Here we focus on animal studies because of issues related to human studies, including experimental invasiveness and infringements of personal privacy, therefore we see limited negative societal impact due to our research. Notably, in the long term, by increasing the efficiency of self-supervised learning techniques, these algorithms can also reduce the amount of data needed, and reduce the number of animals for experiments in neuroscience.

\newpage
\section{Supplementary Figures}

\begin{figure}[h!]
  \centering
  \includegraphics[width=.8\textwidth]{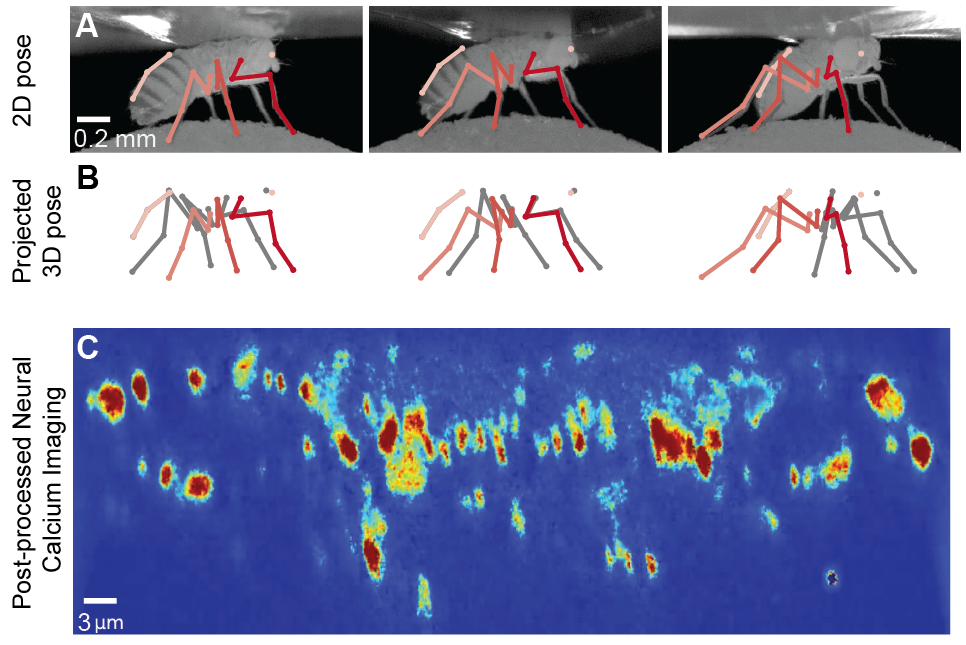}
  \caption{ \textbf{Overview of motion capture and two-photon neural imaging dataset.} A tethered fly (\textit{Drosophila melanogaster}) behaves spontaneously while neural and behavioral data are recorded using multi-view infrared cameras and a two-photon microscope, respectively. The dataset includes \textbf{(A)} 2D poses from six cameras (only three are shown), \textbf{(B)} 3D poses, triangulated from multiview 2D poses. Calibration parameters for the markerless motion capture system are included. \textbf{(C)} Synchronized, registered, and denoised calcium imaging data from coronal sections of the cervical connective. Shown are color-coded activity patterns for populations of descending neurons from the brain (red is active, blue is inactive). Data are collected from multiple animals and include action labels. }
  \label{fig:main}
\end{figure}

\begin{figure}[h!]
  \centering
      \includegraphics[width=13cm]{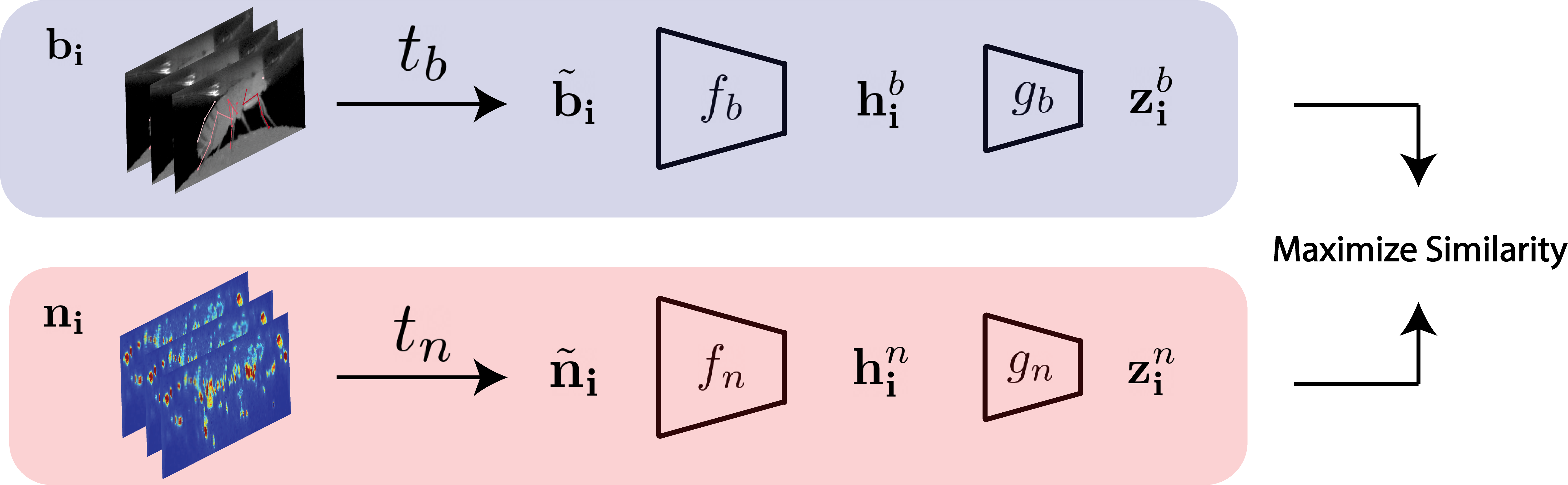}
  \caption{\textbf{Overview of our contrastive learning-based neural action representation learning approach.} First, we sample a synchronized set of behavioral and neural frames, $(\mathbf{b}_{\mathbf{i}}, \mathbf{n}_\mathbf{i})$. Then, we augment these data using randomly sampled augmentation functions $t_b$ and $t_n$. Encoders $f_b$ and $f_n$ then generate intermediate representations $\mathbf{h}^b$ and $\mathbf{h}^n$, which are then projected into $\mathbf{z}_b$ and $\mathbf{z}_t$ by two separate projection heads $g_b$ and $g_n$. We maximize the similarity between the two projections using an InfoNCE loss.}
  \label{fig:method}
\end{figure}

\begin{figure}[h!]%
  \centering
  \includegraphics[width=.8\textwidth]%
  {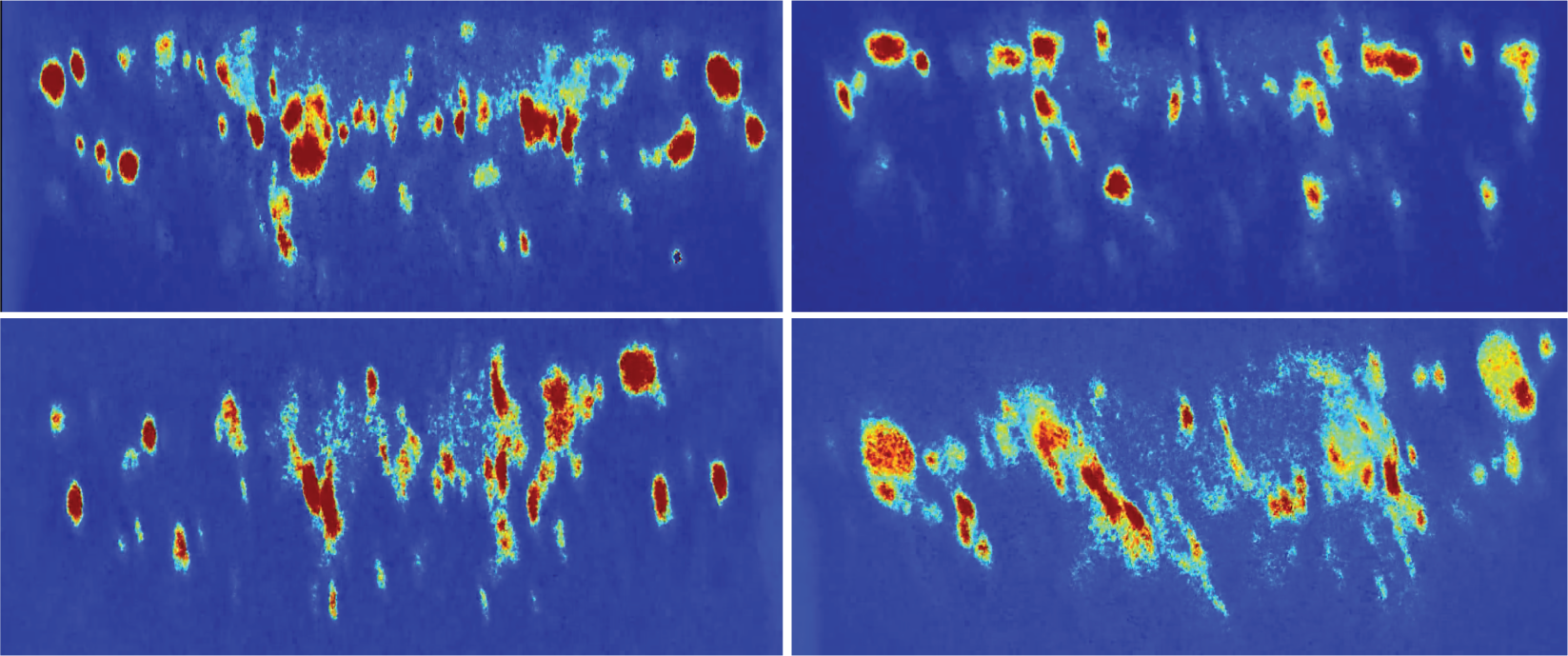}
  \caption{\textbf{Domain gap between nervous systems.} Neural imaging data from four different animals. Images differ in terms of total brightness, the location of observed neurons, the number of visible neurons, and the shape of axons.}
  \label{fig:domain}
\end{figure}

\begin{figure}[h!]
  \centering
  \includegraphics[width=15cm]{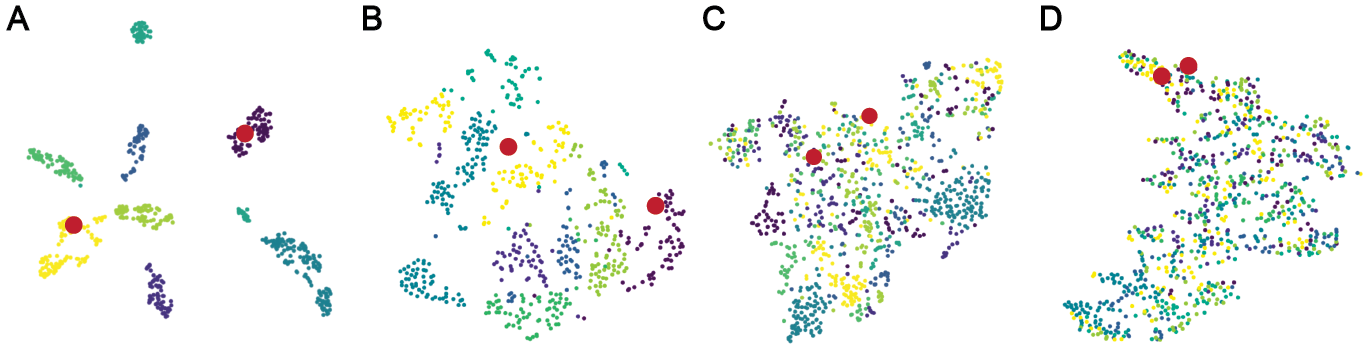}
  \caption{\textbf{t-SNE plots of the neural modality.} Each color denotes a different domain (animal). Two red dots are the embeddings of the same action label. \textbf{(A)} Raw neural data \textbf{(B)} SimCLR representation, \textbf{(C)} Domain adaptation using a two-layer MLP discriminator and a Gradient Reversal Layer \textbf{(D)} Ours, aligns multiple domains and keeps the semantic structure.}
  \label{fig:tsne}
\end{figure}

\begin{figure}[h!]
  \centering
  \includegraphics[width=.85\textwidth]{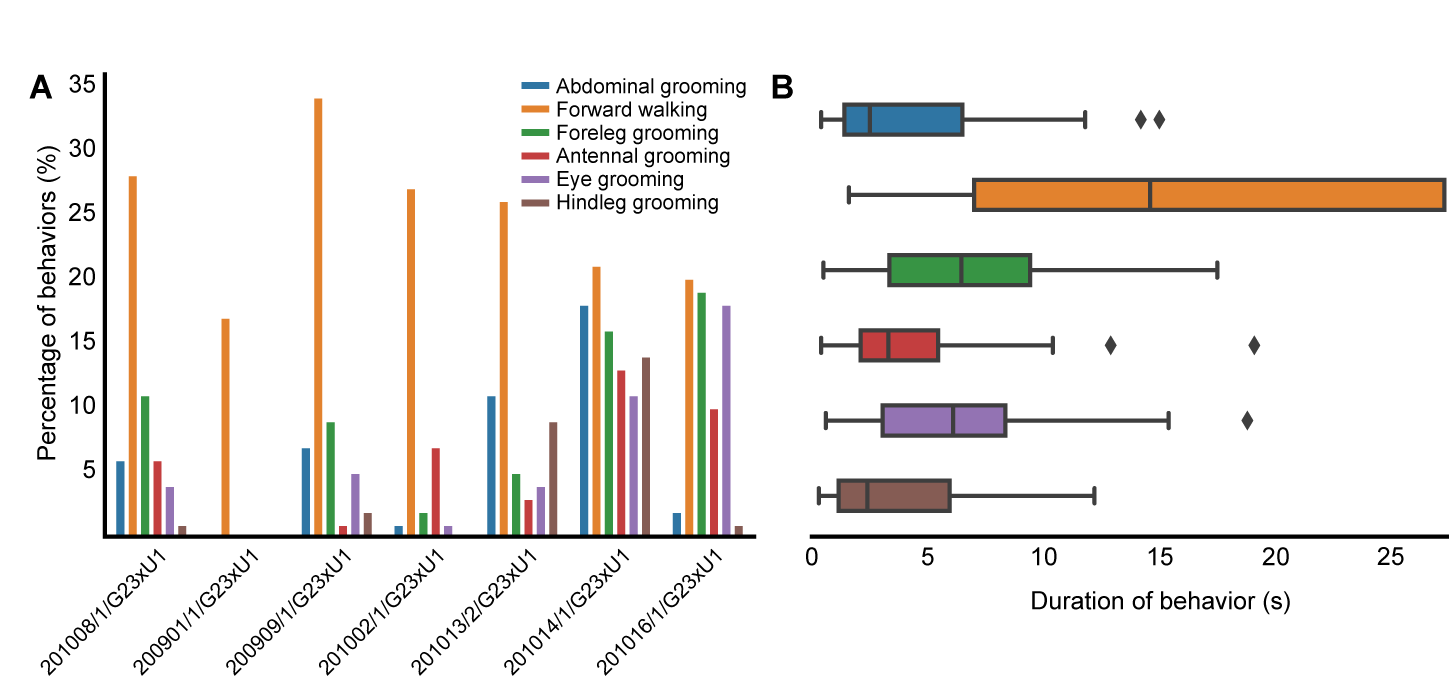}
  \caption{\textbf{Motion Capture and two-photon dataset statistics.} Visualizing \textbf{(A)} the number of annotations per domain and \textbf{(B)} the distribution of the durations of each behavior across domains. Unlike scripted human behaviors, animal behaviors occur spontaneously. The total number of behaviors and their durations do not follow a uniform distribution.}
  \label{fig:dataset}
\end{figure}

\newpage
\section{Supplementary Videos}
\label{sup:video}
The following videos are sample behavioral-neural recordings from two different flies. The videos show \textbf{(left)} raw behavioral RGB video together with \textbf{(right)} registered and denoised neural images in their original resolution. The behavioral video is resampled and synchronized with the neural data. A counter \textbf{(top-left)} shows the time in seconds. The colorbar indicates normalized relative intensity values. Calculation of $\Delta F / F$  values is explained in the Appendix.   \\

\textbf{Video 1:} \url{https://drive.google.com/file/d/1Cepy5xjLj4XiQUITY_yKKu2B4WKdl6nx}

\textbf{Video 2:}
\url{https://drive.google.com/file/d/1OSszc_fMR2Ol2WkUdj1E4u58rFVaMr6E}

\end{document}